\documentclass[11pt]{article}
\usepackage{coling2020}
\usepackage{times}
\usepackage{multirow}

\usepackage{xcolor}

\usepackage{epigraph} 
\usepackage{pdfpages} 
\usepackage{booktabs} 

\usepackage{lineno}

\usepackage[utf8]{inputenc}
\usepackage{fvextra}
\usepackage{csquotes}

\usepackage{amsmath}
\usepackage{graphicx}
\usepackage[colorinlistoftodos]{todonotes} 
\usepackage{listings}
\usepackage{pdfpages}
\usepackage{tcolorbox}
\usepackage{float}
\usepackage{caption}
\usepackage{subcaption}
\usepackage{tikz}
\usetikzlibrary{shapes,arrows}
\usetikzlibrary{matrix}
\usepackage[hyphens]{url}
\usepackage[english]{babel}
\newcounter{quotecount}

\title{Towards Olfactory Information Extraction from Text:\\
A Case Study on Detecting Smell Experiences in Novels}

\author{Ryan Brate
  \and
  Paul Groth \\
  University of Amsterdam \\ Amsterdam, the Netherlands\\
  {\tt r.brate@gmail.com}\\ {\tt p.t.groth@uva.nl} \\ \And 
  Marieke van Erp \\
  KNAW Humanities Cluster \\
  Digital Humanities Lab \\ Amsterdam, the Netherlands \\
  {\tt marieke.van.erp@dh.huc.knaw.nl}}

\date{}
\begin{document}
\maketitle
\begin{abstract}
Environmental factors determine the smells we perceive, but societal factors factors shape the importance, sentiment and biases we give to them. Descriptions of smells in text, or as we call them `smell experiences', offer a window into these factors, but they must first be identified. To the best of our knowledge, no tool exists to extract references to smell experiences from text. In this paper, we present two variations on a semi-supervised approach to identify smell experiences in English literature. The combined set of patterns from both implementations offer significantly better performance than a keyword-based baseline.
\end{abstract}

\section{Introduction}%
\label{sec:introduction}
We rely on our senses: touch, taste, hearing, sight and smell; to complement one another in shaping our interpretation of our environment. There is shifting historical relevance placed on smell - its worthiness for attention, its association with social standing, lifestyle, emotion, science and superstitions, and other topical associations shifting with time \cite{vroon}. English language vocabulary specific to the description of smell experiences is not expansive, and, to the best of our knowledge, language technology to identify references to smell in text even less so. A topic search in the Cambridge Dictionary online of words categorised as relating to smells and smelling\footnote{\url{https://dictionary.cambridge.org/topics/senses-and-sounds/smells-and-smelling/} Accessed: 5 August 2020} returns fewer than 30 words, that are predominantly concerned with \textit{intensity} or \textit{sentiment} such as \textit{fetid} and \textit{reek}. Other characteristics of smell are instead often described in terms of reference smell sources as similes such as \textit{There is a strange unwholesome smell upon the room, \textbf{like mildewed corduroys}}.


In this paper, we present a dataset of annotated references to smell, which we call `smell experiences' in literature, as well as a first approach and experiments to automatically recognise references to smells in texts. 

The remainder of this paper is organised as follows. In Section~\ref{relwork}, we discuss related work. In Section~\ref{sub:datasets}, we describe our corpus and its creation process. In Section~\ref{sec:experiments}, we present our extraction approach and experiments. In Section~\ref{sec:eval}, we discuss the results. We conclude with Section~\ref{sec:conc} in which we present our conclusions and directions for future work. Our data is available at \url{ http://doi.org/10.5281/zenodo.4199996} and the code to the experiments is available at \url{https://github.com/DHLab-nl/Detecting-Smell-Experiences-in-Novels}. This work is a preliminary result of the Odeuropa project which will commence formally in January 2021: \url{https://odeuropa.eu/}.

\section{Related Work}\label{relwork}
The cultural significance of smells is a niche topic in the humanities domain, but one that has recently gained more interest with a translation of Muchembled's 2017 \textit{La Civilisation des odeurs (XVIe si\`{e}cle-d\'{e}but XIXe si\`{e}cle)} to English \cite{muchembled} and Barwich's \textit{Smellosophy: what the nose tells the mind}\cite{barwich2020smellosophy} being reviewed in mainstream media.\footnote{cf. \url{https://www.spectator.co.uk/article/where-are-the-scents-of-yesterday-entire-countries-have-lost-their-distinctive-smell} ; \url{https://slate.com/culture/2020/07/smells-history-book-review-france-plague-farts.html} ; \url{https://www.wsj.com/articles/smells-and-smellosophy-review-what-the-nose-knows-11594391739}} While historians interested in olfaction such as \cite{tullett2019smell} analyse textual accounts of experienced smells, (computational) linguistic analysis of smell experiences (at least for English) has received little attention. 

This lack of attention may be due to the fact that Western languages such as English and Dutch do not contain rich vocabularies for describing odorants as opposed to some other languages. In \cite{majid2014odors} Jahai and English speakers and in \cite{majid2018olfactory} Jahai and Dutch speakers were contrasted in describing a range of odorants. The Jahai, a group of nomadic hunter-gatherers in Malaysia, have over a dozen terms to describe odours. In the experiment, the Jahai speakers were both more consistent and greatly more controlled in the terms they used than the English and Dutch speakers relying on reference smell sources.  

There is evidence that consistency in the use of smell sources in English can be conditioned. \newcite{Croijmans_majid} examined the accuracy and consistency of wine experts, coffee experts and people with no expertise in identifying smells. No group was better at naming smells outside of the domain of their expertise. However, it was apparent the domain experts had developed a toolkit of common smell sources they frequently drew upon. An experiment on predicting properties of wines from experts' wine reviews confirms this~\cite{hendrickx-etal-2016-quaffable}. 

Sensorial lexicons have been developed that include terms related to smell (cf. \newcite{tekiroglu-etal-2014-computational}). However, as the initial seed words to bootstrap the lexicon for smell are limited, the olfactory clusters in such lexicons are by extension also limited. A different approach is taken in \cite{kiela-etal-2015-grounding}, where terms related to smells are connected to their chemical compounds. While this is useful for translating olfactory information from the chemistry domain to language, it does not aid us in recognising the wide variety of expressions used in texts to describe smell experiences.  


\section{Literary Smell Dataset}
\label{sub:datasets}
To begin to tackle the problem of recognising smell experiences, we created a unique dataset focused on such expressions in literary texts. Specifically, by selecting texts from Project Gutenberg\footnote{\url{https://www.gutenberg.org/}} that had the highest rate of occurrence of keywords derived from Table~\ref{tab:cambridge}, we assembled a set 139 English literary texts. Each sentence in this collection was tokenised using NLTK\footnote{\url{https://nltk.org}}, and POS-tagged and syntactically parsed using spaCy~\cite{spacy}.\footnote{\url{https://spacy.io/}} We split the set into three datasets: a harvesting dataset of 99 texts; a validation dataset of 20 texts; and an evaluation dataset of 20 texts.  

\begin{table}[htp]
    \footnotesize
    \centering

    \begin{tabular}{|c|c|c|}
        \hline
        \textbf{smell-only} & \textbf{smell-only} & \textbf{smell and taste} \\
        \textbf{(in all contexts)} & \textbf{(in sensory contexts)} & \textbf{only} \\
        \hline
        \textit{odour(N)}, \textit{odorous(A)} & \textit{fragrance(N)}& \underline{pungent(A)} \\
        \textit{malodorous(A)} & \textbf{musk(N)}  & \underline{pungency(N)}\\
        \underline{\textbf{\textit{fetid(A)}}}, \underline{\textbf{\textit{foetid(A)}}} & \textbf{\textit{fusty(A)}},\textit{frowsty(A)} & \underline{pungently(ADV)} \\
        \textit{whiffy(A)} &\underline{\textit{ripe(A)}}, \underline{\textit{ripeness(N)}}& \textit{savour(N,V)}\\
        smell(N,V), \textit{scent(N)}& \underline{\textit{reek(N,V)}}, \textit{stink(N,V)}& \textit{\underline{\textbf{acrid(A)}}}\\
        \textit{smelly(A)} & \textit{stench(N)}, \textit{niff(N)} & \\
        \underline{\textit{scented(A)}} & sniff(V), \textbf{piney(A)} & \\
        \textit{perfume(N)} & waft(N,V), \textit{stinky(A)} &\\
        \underline{\textit{aroma(N)}}, \textit{aromatic(N)} &  \underline{whiff(N)}, &\\
        \textit{fragranced(A)} &&\\
        \textit{\textbf{petrichor(N)}} &&\\
        \textit{\textbf{musty(A)}}, \textbf{musky(A)} &&\\
        \hline
         
    \end{tabular}
    \text{Note 1: A,N,V denotes adjectives, nouns and verbs, respectively}\\
    \text{Note 2: \underline{underlined}: words with smell strength connotations}\\
    \text{Note 3: \textit{italicised}: words with sentiment associations}\\
    \text{Note 4: \textbf{bold face}: describes characteristics beyond strength or sentiment}\\
    \caption{Results from Cambridge Dictionary `smells and smelling' SMART Thesaurus search \\}
 \label{tab:cambridge}
\end{table}

From the evaluation dataset, a gold standard of manually labelled sentences was assembled consisting of seven documents, each of 100 randomly assigned sentences from multiple literary texts and annotated by a single annotator. To evaluate the inter-annotator agreement between the three annotators, one additional document consisting of 100 randomly selected sentences was annotated independently by the annotators.   

Despite having chosen the harvesting, validation and extract sets for their high frequency of Table~\ref{tab:cambridge} derived keywords, on average only approximately 1 in 100 sentences contain a keyword. Thus, assuming that smell experiences typically contain a keyword, the evaluation set contains smell experiences in very low proportion. A gold standard set of extracts was sampled from the evaluation dataset to ensure a substantial number of smell extracts. The evaluation set was scanned for the high smell association keywords derived from Table~\ref{tab:cambridge}.  80\% of the sentences in the gold standard documents contain a Table~\ref{tab:cambridge} related word, the remaining 20\% were randomly sampled. There is no overlap between documents, or redundancy within a document.

Annotators were asked to highlight and annotated spans according to the following criteria:\footnote{The full annotation guidelines can be found on our Github page}

\begin{itemize}
    \item `d'. A smell description;
        e.g., `\underline{An odd fragrance, a smell of damp plaster}, wafted from the new house to his senses'.\\
        The inherent subjectivity of when precisely a smell experience becomes a description is left to the perception of the annotator.
    \item `o'. A smell alluded to without expansion of its characteristics;
        e.g., `\underline{A fragrance} wafted from the new house to his senses'.
    \item `v'. Any verb in the sentence which is associated with smell generally or with a specific smell experience within the extract;
        e.g., `An odd fragrance \underline{wafted from} the new house to his senses';
    \item `s'. Sense of smell alluded to directly;
        e.g., `An odd fragrance, wafted from the new house to \underline{his senses}'.
\end{itemize}

Additionally, two documents (of the aforementioned group of 7) were annotated with an additional set of tags:
\begin{itemize}
    \item `a'. An adjective being applied to the smell alluded to;
        e.g., `An \underline{odd} fragrance, a smell of damp plaster, wafted from the new house to his senses'.
    \item `n'. The noun group referred to as a smell source;
        E.g., `An odd fragrance, a smell of \underline{damp plaster}, wafted from the new house to his senses'.
\end{itemize}

\begin{table}[t]
    \footnotesize
    \centering
    \begin{tabular}{|c|c|}
        \hline
        Annotation tag & Number of corresponding text spans\\
        \hline
        `d' & 533\\
        \hline
         o' & 129\\
        \hline
        `v' & 186\\
        \hline
        `s' & 34\\
        \hline
        `a' & 37\\
        \hline
        `n' & 75\\
        \hline
    \end{tabular}
    \caption{Number of text spans by annotation tag}
    \label{tab:tags}
\end{table}

 The inter-annotator agreement is measured using Cohen's Kappa~\cite{cohen1960kappa} per single gold standard document in a pairwise fashion. Specifically:
        \begin{itemize}
            \item The level of agreement with respect to those sentences which were tagged as \textit{a smell experience}, by one or more annotators, i.e., those sentences with a text span annotated with either `d' or `o'.
            \item The level of agreement with respect to those sentences which were tagged as \textit{a smell description}, by one or more annotators, i.e., those sentences with a text span annotated with `d'.
            \item The level of agreement of those verbs within sentences what were tagged with `v', i.e., verbs the annotator perceives as being related to smell generally, or in the context of the sentence.
        \end{itemize}

\begin{figure}[t]
    \centering
    \includegraphics[width=0.8\textwidth]{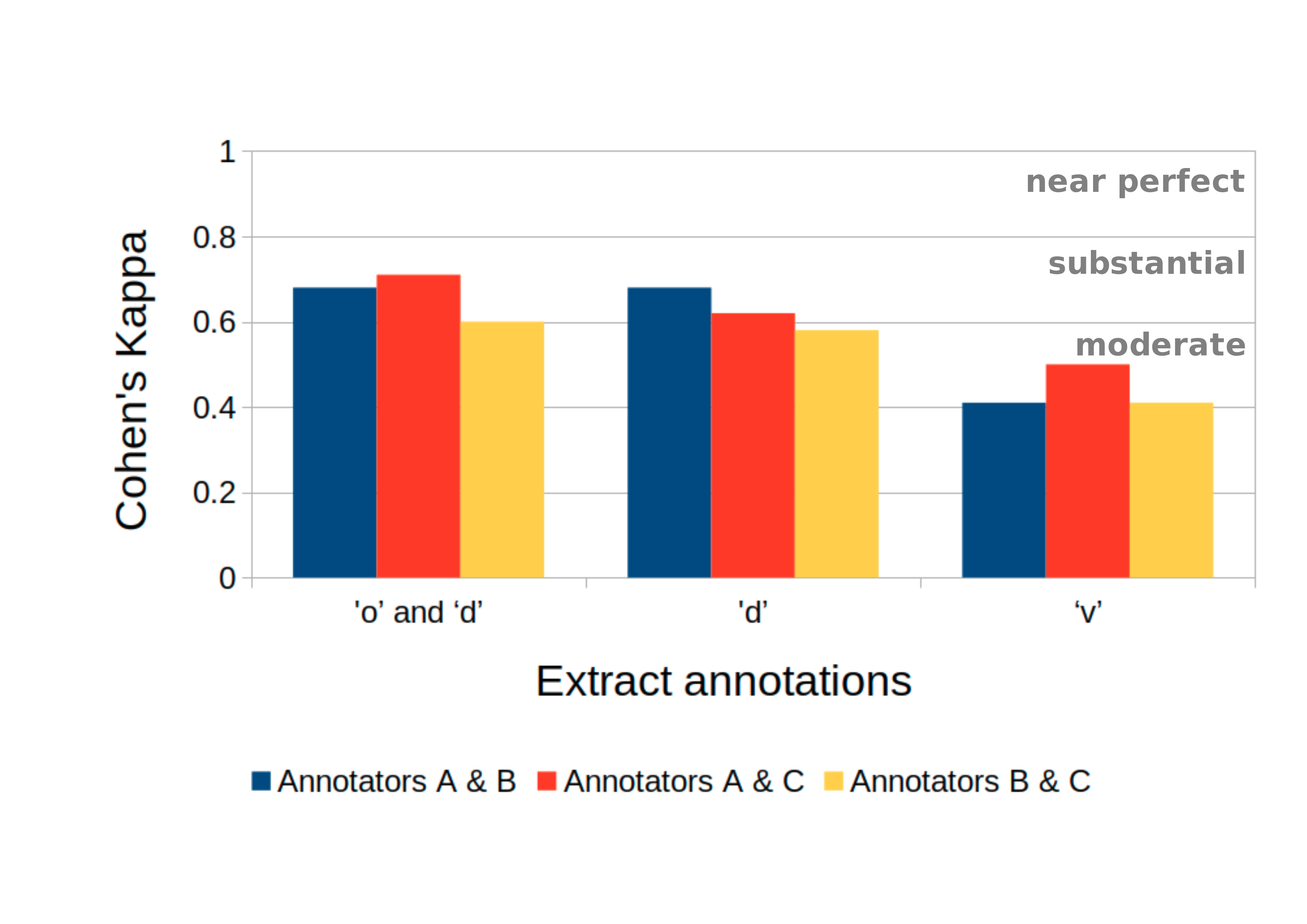}
    \caption{Cohen’s Kappa scores of pairwise annotator agreement}
    \label{fig:cohens_bar}
\end{figure}

Cohen's Kappa is a metric used to measure pairwise inter-annotator agreement. A Cohen's Kappa of 0, denotes an even probability of agreement. \newcite{Landis} denote a Cohen's Kappa score of 0.41 to 0.60, and .61 - 0.80 as representing \textit{moderate} and \textit{substantial} strength of agreement, respectively. A score of 0.81 to 1.0 can be considered as near perfect agreement. 

Figure~\ref{fig:cohens_bar} shows the pairwise annotator agreement with regards the single gold standard document of 100 extracts, annotated by multiple annotators. All annotators are in \textit{substantial} agreement in identifying all and any extracts that allude to smell, i.e., all spans labelled `o' or `d'. Annotators are generally in substantial agreement in identifying extracts which \textit{describe} smell experiences, i.e., extracts with a spans labelled `d'. Although, one pair of annotators are at the very upper end of \textit{moderate} agreement only. Finally, in identifying verbs either highly associated with smell, or associated with smell in the context, there was only \textit{moderate} annotator agreement.

It is reasonable to assume that human error, i.e., misreading, miscomprehending or simply skipping an extract, played some role in the observed imperfect inter-annotator agreement scores. Instances of likely human error are apparent on inspection of the gold standard, in those instances where there is arguably little room for personal subjectivity, for example, one of the three annotators did not attribute either a `d' or `o' tag to the span: \textit{There was a smell of decaying leaves and of dog.}

However, a number of extracts clearly demonstrate the potential for subjectivity in smell experience interpretation, as a source for annotator disagreement, for instance, in the following extract, each of the three annotators attributed `d', `o' and no tag to it, respectively: \textit{Seated beside her aromatic rest, In silence musing on her loveliness, Her knight and troubadour.}

In the following extract, one of three annotators tagged it as `o', the other two presumably thought it sufficiently descriptive to be tagged `d': \textit{Between each pair of columns an elegant table of cedar bore on its platform a bronze cup filled with scented oil, from which the cotton wicks drew an odoriferous light.}

\newpage
\section{Extraction Approaches}\label{sec:experiments}
As there is no known corpus annotated with smell expressions available to train a supervised smell language recogniser, we investigate pattern-based approaches to recognising such expressions. 
We base our approach on the concepts put forward in approaches such as the detection of hypernym-hyponym pairs~\cite{Hearst}, e.g., \textit{car is a type of vehicle}, and iterative bootstrapping~\cite{brin} to detect author-book title pairs. 

We start with seed features that are matched against the harvesting dataset. The resulting matches are then manually evaluated to identify new linguistic pattens, which are in-turn used to identify new features for the next bootstrapping round.
The process can be seeded by introducing known features into the lexicon, or known patterns into the pattern set. In both~\cite{Hearst} and~\cite{brin}, this process was used to assemble a lexicon, the pattern set being effectively a by-product of the process. However, for our purposes, it is the set of patterns that is of interest, and their potential use to identify smell experiences. 

The textual features targeted in prior work represented distinct real world concepts linked through a conceptual relationships. The expression of smell (in English) does not conform to a natural set of paired entities reflecting a relationship in the same way as authors and book titles do. However, surrounding adjectives, and verb and noun groups help characterise the smell experience and can therefore be targeted as complements. 
 
A difficulty of identifying smell references over hypernym-hyponym or author-book title pairs is that there is no inherent relationship defining the number of coincident complements necessary.  Instead it is a question of how restrictive we wish to make the criteria for matching sentences in the harvesting dataset. The greater the number of complements, the fewer extracts we can expect to retrieve. Thus, a too restrictive choice may result in stalling bootstrapping process. Conversely, if the choice is not restrictive enough and too many extracts unrelated to smell are returned, the process becomes uninformative. In evaluating smell related vocabulary, \cite{iatropoulos} concluded that the most commonplace words used in smell descriptions are those that could apply to a wide range of sensory contexts. This makes intuitive sense given the heavy reliance on reference smell sources to define smell characteristics. Hence, single complements are not targeted, as being too relevant outside of smell contexts. In this pilot project, we therefore focused on at least two complements in a pattern.

A basic assumption in our approaches is that found complements are indicative of the presence of a smell expression. We therefore aim to detect the following types of smell expression complements: 
\begin{enumerate}
    \item \textbf{Adjectives} modifying the smell experience perception, and the coincident \textbf{noun group} acting as the reference smell (one or more nouns modified by adjectives); E.g., \textit{`An \textbf{odd} fragrance, a smell of \textbf{damp plaster}, wafted from the new house to his senses'},where `odd' and `damp plaster' are the adjective and noun group, respectively. Thus, it is assumed that `odd' and `damp plaster', being coincident in defining this smell experience, are indicative of a smell experience when both present in other extracts. 
    \item A \textbf{noun group} acting as the reference smell, and the coincident \textbf{verb group} (adverbs, verbs, associated prepositions) describing how the smell moves; E.g., `An odd fragrance, a smell of \textbf{damp plaster}, \textbf{wafted from} the new house to his senses', where `damp plaster' and `wafted from' are the noun group and verb group, respectively. Thus, it is assumed that `damp plaster' and `wafted from', being coincident in defining this smell experience, are indicative of a smell experience when both present in other extracts.
\end{enumerate}

\subsection{Approach 1: Targeting Adjective and Noun Groups}
To capture and enable pattern matching with parts of speech, synonyms and flexible groupings of these in text, we start with words derived from Table~\ref{tab:cambridge} assembled in synonym groups. For example, \textit{$<$smell\_noun$>$} is defined to match against `aroma', `odour', `scent', `perfume' etc. The patterns further contain part of speech chunks that can match various tokens. These chunks were defined, and updated in process, based on observed language patterns from harvesting set extracts. Listing~\ref{lst:examplepattern} is an example of a high-level pattern representation.

\begin{lstlisting}[caption={Example identified pattern}, label=lst:examplepattern,captionpos=b]
[<adj>] <smell_noun> _,_* _of_ <pronoun>* [<noun> {_of_ <noun>}*]
\end{lstlisting}

 It will match the boldfaced adjective and noun groups in for example \textit{the \textbf{warm} aroma of \textbf{multitudinous exotics}} and \textit{the \textbf{ammoniacal} smell of \textbf{the horses}}

\subsection{Approach 2: Targeting Verb and Noun Groups}
As in the Adjectives \& Nouns approach, we start with the same high-smell association seed lexicon entry, \textit{\_aroma\_NOUN}, as it results in a very manageable number of bootstrapped extracts. Based on the resulting extracts, new patterns are hypothesised: e.g., from the extract \textit{`the aroma of the \textbf{newly-sawn timber and saw dust} \textbf{mingled} in the air'}, we may hypothesise the pattern \textit{$<$smell\_noun$>$ \_of$|$like\_ \_\_DET* $<$pronoun$>$* \textbf{[$<$noun$>$ \{\_of\_ $<$noun$>$\}*]} \textbf{[$<$verb$>$ prep\_\_*}]}. 

This approach also includes a validation loop. As patterns with low precision risk introducing a large volume of vocabulary into the lexicon which is unrelated to smell. The validation loop is an attempt to limit this, by estimating pattern precision and setting a minimum acceptance threshold by taking 10 extracts per pattern and manually tagging them as \textit{true positive}, \textit{false positive}, or \textit{unknown}. Patterns that pass a validation threshold of 0.7 estimated precision are accepted. If no example of a pattern is present in the validation set, it is removed.

Two variations of the patterns are retained, \textit{identification patterns} and \textit{extraction patterns}. \underline{Extraction patterns} target the previously discussed feature pairs, such to introduce new vocabulary into the lexicon. Thus, extraction patterns are used to drive each iterative cycle. For example, based on the preceding hypothesised pattern, e.g., \textit{$<$smell\_noun$>$ \_of$|$like\_ \_\_DET* $<$pronoun$>$* [$<$noun$>$ \{\_of\_ $<$noun$>$\}*] [$<$verb$>$ prep\_\_*]}. \underline{Identification patterns} are a superset of the extraction pattern set, and are concerned with matching any and all smell experiences, not just matching feature pairs. Several identification patterns my be derived from a single hypothesised extraction pattern, expressing the potential variation in matching smell experiences. The identification pattern set is our desired output from the iterative bootstrapping process. In the final step of the approach, the extraction patterns are applied to the harvesting set, and targeted complements are collected and added to the lexicon. 

\section{Evaluation and Discussion}\label{sec:eval}
Four complete cycles of the Adjectives \& Nouns approach and three complete cycles of the Verbs \& Nouns approach were performed. 

The Adjectives \& Nouns approach identified 48 new identification patterns as seen in Table \ref{tab:implementation1}. The  majority of these patterns involve Table~\ref{tab:cambridge} derived words such as \textit{$<$adj$>$* compound\_\_ $<$smell\_noun$>$} matching \textit{a delightful forest \textbf{aroma}}, and \textit{$<$adj$>$ \_with\_ \_\_DET* $<$pronoun$>$* $<$smell\_noun$>$ \_of\_ $<$pronoun$>$* $<$verb$>$ $<$noun$>$ \{\_of\_ $<$noun$>$\}*'} matching  \textit{heavy with the \textbf{smell} of freshly turned soil}. 

Additionally, a small number of patterns identified do not involve the Table~\ref{tab:cambridge} vocabulary, such as \textit{$<$adj$>$* \_breath$|$breaths\_ \_of\_ $<$pronoun$>$* $<$noun$>$ \{\_of\_ $<$noun$>$\}*} matching \textit{`...and inhale the sweet breath of autumn, which was borne upon gentle gales'} and \textit{`\_air\_* \_,\_ \_sweet\_ \_with\_ $<$pronoun$>$* $<$noun$>$ \{\_of\_ $<$noun$>$\}*'} matching \textit{ `the mild air, sweet with fading leaves and bracken'}. 

The Verbs \& Nouns approach identified 31 new identification patterns as show in Table \ref{tab:implementation2}. Again a majority involve Table~\ref{tab:cambridge} derived words, for example \textit{`$<$smell\_noun$>$ \_of|like\_ \_\_DET* $<$pronoun$>$* \textbf{$<$noun$>$ \{\_of\_ $<$noun$>$\}*} \textbf{$<$verb$>$ prep\_\_}'} matching phrases such as \textit{the aroma of \textbf{new-sawn timber and sawdust} \textbf{mingled  with}...'}. The single example of lexico-syntactic pattern not involving Table~\ref{tab:cambridge} derived words, introduced the \textit{`incense'} as synonymous with smell: \textit{`\_fumes\_ \_of\_ \_incense\_ \{\_of\_ $<$noun$>$\}* \_\_DET* $<$verb$>$ prep\_\_*'} matching for example \textit{`the heavy fumes of incense rose up'}

\begin{table}[h]
    \footnotesize
    \centering
    
    \begin{tabular}{|c|c|c|c|c|}
    \hline 
    Cycle & Lexicon & New (unseen) & Hypothesised &  New id. patterns/ \\
          & entries & extracts   &  (new) patterns             & New ex. patterns\\
    \hline
    0 & 1**  & \textbf{91} & 15  & 15 / 13\\
    \hline
    1 & 519 & \textbf{1,509} & 28  & 26 / 22\\ 
    \hline
    2*** & 874 & \textbf{4,216} & 14  & 13 / 8\\
    \hline 
    3 & 463 & \textbf{464}  & 4 **** & 4 / 4 \\
    \hline
    \end{tabular}\\
    \text{**Seed word: \textit{\_aroma\_NOUN} }\\
    \text{*** sifted with Table~\ref{tab:cambridge} word search due to high volume} \\
    \text{**** not subject to validation as cycles stopped}\\
    \text{Note 1:  Each lexicon (pair) entry is unique, and each extract is unique}\\
    \caption{Record of iterative cycles outcomes for Approach 1: targeting coincident adjectives modifying the smell, and noun group reference smells}
    \label{tab:implementation1}
\end{table}

\begin{table}[h]
    \footnotesize
    \centering
    
    \begin{tabular}{|c|c|c|c|c|}
    \hline 
    Cycle & Lexicon & New (unseen) & Hypothesised &  New id. patterns/ \\
          & entries & extracts   &  (new) patterns             & New ex. patterns\\
    \hline
    0 & 1**  & \textbf{91} & 11 & 10 / 9\\
    \hline
    1*** & 530 & \textbf{2,968} & 12 & 10 / 9\\ 
    \hline
    2 & 565 & \textbf{1,030} & 11 ****& 11 / 8\\
    \hline 
    \end{tabular}\\
    \text{**Seed word: \textit{\_aroma\_NOUN} }\\
    \text{*** sifted with Table~\ref{tab:cambridge} word search due to high volume} \\
    \text{**** not subject to validation as cycles stopped}\\
    \text{Note: Each lexicon (pair) entry is unique, and each extract is unique}
\caption{Record of iterative cycles outcomes for Approach 2: targeting coincident verb groups associated with the smell experience and noun group reference smells}
    \label{tab:implementation2}
\end{table}

\begin{figure}[t]
    \centering
    \includegraphics[width=.9\linewidth]{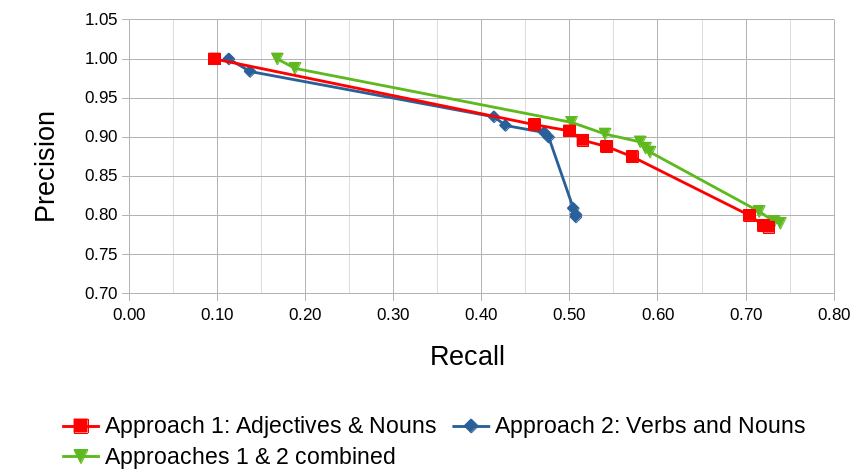}
    \caption{Group prediction precision-recall performance of the pattern sets of: approach 1; approach 2; and approach 1 and 2 combined.}%
    \label{fig:precison_recall_B}
\end{figure}

Figure~\ref{fig:precison_recall_B} shows the relative precision-recall performance of the group predictions, in respect of the gold standard, with regards the pattern sets of: approach 1; approach 2; and both combined. 

It is apparent that the Adjectives \& Nouns approach (red) consistently outperformed the pattern set of Verbs \& Nouns approach (black); and that they target complementary smell expressions (combined approach, green). The relative performance of the combined approach with that of the Verbs \& Nouns approach. The combined approach is significantly better with a 5\% significance level where patterns have a precision greater than 0.75. The combined pattern set \textit{significantly} has superior recall performance at corresponding precision cut-offs. Thus, we can conclude that targeting different feature pairs did result in patterns that target different smell extracts in the gold standard set.

\section{Conclusions and Future Work}\label{sec:conc}
In this paper, we set out to identify smell experiences in English literary texts. We created a gold standard dataset and our experiments demonstrated that iterative bootstrapping techniques can be used to identify smell experiences in text. 

Whilst the overwhelming majority of identified patterns involved keywords and phrases with a high smell association, the implementations revealed a number of new phrases used in smell contexts. Furthermore, we showed that at the very highest levels of precision, pattern group identification of smell experiences offers significantly better recall rates than a keyword search. 

The focus of application of iterative bootstrapping implementations was centred on single sentence extracts for English literary texts. It would be interesting to explore the applicability of this semi-supervised method to other textual contexts, and longer-distance relationships spanning multiple sentences. Additionally, it would be informative to explore the influence of tweaking the implementation's parameters and approaches, including the pre-processing steps such as instead of using a shallow parser and experimenting with dependency grammars or semantic role labelling. With regards the precision-recall performance of the resulting pattern sets, such as exploring the influence of different seed words on the process outcomes, investigate the impact of a higher validation precision threshold for the number, and quality, of extracts returned each cycle and the corresponding identified patterns.

Our experiments demonstrated the potential use of patterns for identifying textual smell experiences in text. However, the number of extracts, and the quality of extracts in terms of smell experience density was identified as source of inefficiency which would benefit from being addressed further. The statistics presented in Table~\ref{tab:implementation1}, for example, show the explosion in the number extracts for manual examination, in certain cycles, which correspond to only comparatively few new patterns being identified.

The observed high volume of low smell experience extracts may be an inherent challenge of smell experiences relying on vocabulary which is equally, or more so, applicable in other sensory contexts. I.e., the targeting and adding to the lexicon of words with a low smell association, resulting in poorer quality extracts. However, there are a number of clear, possible avenues to explore such as increasing the validation set size, ensuring more accurate precision estimates thus improving the level of smell association of lexicon entries on average and exploring the effects of using more coincident features simultaneously, i.e., pairs were selected on the basis that if one feature alone was weakly associated with smell, two together may improve the association. More coincident features may further improve the likelihood of an extract relating to smell.

Finally, consideration as to degree of agreement between people in their interpretation, suggested a less than perfect agreement not only of the subtly nuanced aspects of smell experiences, but even at recognition of smell experiences as a broad classification. On inspection of annotations, however, it is unclear how many of these were genuine discrepancies in terms of subjective perceptions. More annotators, supported by a more comprehensive approach to tagging, e.g., requiring the annotators to explicitly note their reasoning and deliberations would help. This would offer a window into the mind of the annotators, reinforcing any conclusions that may be drawn. 

Whilst there is extensive scope for further study, the experiments in this paper have shown that smell experiences can be identified, and smell experience features can be extracted from text providing a useful foundation for understanding smell usage within literature. 

 \section*{Acknowledgements}
The authors would like to thank the annotators for taking the time to think about smelly language and tag multiple sets of sentences for us. 

\clearpage
\bibliographystyle{coling}
\bibliography{coling2020}

\begin{thebibliography}{}

\bibitem[\protect\citename{Barwich}2020]{barwich2020smellosophy}
Ann-Sophie Barwich.
\newblock 2020.
\newblock {\em Smellosophy: What the Nose tells the Mind}.
\newblock Harvard University Press.

\bibitem[\protect\citename{Brin}1998]{brin}
Sergey Brin.
\newblock 1998.
\newblock Extracting patterns and relations from the world wide web.
\newblock In {\em The World Wide Web and Databases (WebDB 1998)}, 06.

\bibitem[\protect\citename{Cohen}1960]{cohen1960kappa}
Jacob Cohen.
\newblock 1960.
\newblock Kappa: Coefficient of concordance.
\newblock {\em Educational and Psychological Measurement}, 20(37).

\bibitem[\protect\citename{Croijmans and Majid}2016]{Croijmans_majid}
Ilja Croijmans and Asifa Majid.
\newblock 2016.
\newblock Not all flavor expertise is equal: The language of wine and coffee
  experts.
\newblock {\em PLoS ONE}, 11(6):e0155845.

\bibitem[\protect\citename{Hearst}2000]{Hearst}
Marti Hearst.
\newblock 2000.
\newblock Automatic acquisition of hyponyms from large text corpora.
\newblock {\em Proceedings of the 14th Conference on Computational Linguistics
  (CoLing)}, 05.

\bibitem[\protect\citename{Hendrickx \bgroup et al.\egroup
  }2016]{hendrickx-etal-2016-quaffable}
Iris Hendrickx, Els Lefever, Ilja Croijmans, Asifa Majid, and Antal van~den
  Bosch.
\newblock 2016.
\newblock Very quaffable and great fun: Applying {NLP} to wine reviews.
\newblock In {\em Proceedings of the 54th Annual Meeting of the Association for
  Computational Linguistics (Volume 2: Short Papers)}, pages 306--312, Berlin,
  Germany, August. Association for Computational Linguistics.

\bibitem[\protect\citename{Iatropoulos \bgroup et al.\egroup
  }2018]{iatropoulos}
Georgios Iatropoulos, Pawel Herman, Anders Lansner, Jussi Karlgren, Maria
  Larsson, and Jonas~K. Olofsson.
\newblock 2018.
\newblock The language of smell: Connecting linguistic and psychophysical
  properties of odor descriptors.
\newblock {\em Cognition}, 178:37 -- 49.

\bibitem[\protect\citename{Kiela \bgroup et al.\egroup
  }2015]{kiela-etal-2015-grounding}
Douwe Kiela, Luana Bulat, and Stephen Clark.
\newblock 2015.
\newblock Grounding semantics in olfactory perception.
\newblock In {\em Proceedings of the 53rd Annual Meeting of the Association for
  Computational Linguistics and the 7th International Joint Conference on
  Natural Language Processing (Volume 2: Short Papers)}, pages 231--236,
  Beijing, China, July. Association for Computational Linguistics.

\bibitem[\protect\citename{Landis and Koch}1977]{Landis}
J.~Richard Landis and Gary~G. Koch.
\newblock 1977.
\newblock The measurement of observer agreement for categorical data.
\newblock {\em Biometrics}, 33(1):159--174.

\bibitem[\protect\citename{Majid and Burenhult}2014]{majid2014odors}
Asifa Majid and Niclas Burenhult.
\newblock 2014.
\newblock Odors are expressible in language, as long as you speak the right
  language.
\newblock {\em Cognition}, 130(2):266--270.

\bibitem[\protect\citename{Majid \bgroup et al.\egroup
  }2018]{majid2018olfactory}
Asifa Majid, Niclas Burenhult, Marcus Stensmyr, Josje De~Valk, and Bill~S
  Hansson.
\newblock 2018.
\newblock Olfactory language and abstraction across cultures.
\newblock {\em Philosophical Transactions of the Royal Society B: Biological
  Sciences}, 373(1752):20170139.

\bibitem[\protect\citename{Montani \bgroup et al.\egroup }2020]{spacy}
Ines Montani, Matthew Honnibal, Matthew Honnibal, Sofie~Van Landeghem, Henning
  Peters, Adriane Boyd, Maxim Samsonov, Jim Geovedi, Jim Regan, Gy\"{o}rgy
  Orosz, Paul~O'Leary McCann, S{\o}ren~Lind Kristiansen, Duygu Altinok, Roman,
  Leander Fiedler, Gr\'{e}gory Howard, Explosion Bot, Sam Bozek, Wannaphong
  Phatthiyaphaibun, Mark Amery, Bj\'{o}rn B\"{o}ing, Pradeep~Kumar Tippa, Yohei
  Tamura, Leif~Uwe Vogelsang, Ramanan Balakrishnan, Vadim Mazaev, GregDubbin,
  jeannefukumaru, Jens~Dahl M{\o}llerh{\o}j, and Avadh Patel.
\newblock 2020.
\newblock {explosion/spaCy: v2.3.2: Improved Korean tokenizer speed,
  experimental character-based pretraining and bug fixes}, July.

\bibitem[\protect\citename{Muchembled}2020]{muchembled}
Robert Muchembled.
\newblock 2020.
\newblock {\em Smells: A Cultural History of Odours in Early Modern Times}.
\newblock Polity.

\bibitem[\protect\citename{Tekiro{\u{g}}lu \bgroup et al.\egroup
  }2014]{tekiroglu-etal-2014-computational}
Serra~Sinem Tekiro{\u{g}}lu, G{\"o}zde {\"O}zbal, and Carlo Strapparava.
\newblock 2014.
\newblock A computational approach to generate a sensorial lexicon.
\newblock In {\em Proceedings of the 4th Workshop on Cognitive Aspects of the
  Lexicon ({C}og{AL}ex)}, pages 114--125, Dublin, Ireland, August. Association
  for Computational Linguistics and Dublin City University.

\bibitem[\protect\citename{Tullett}2019]{tullett2019smell}
William Tullett.
\newblock 2019.
\newblock {\em Smell in Eighteenth-century England: A Social Sense}.
\newblock Oxford University Press.

\bibitem[\protect\citename{Vroon \bgroup et al.\egroup }1997]{vroon}
Piet Vroon, Anton Amerongen, Hans, and Vries.
\newblock 1997.
\newblock {\em Smell: The secret seducer}.
\newblock Farrar, Straus and Giroux.

\end{thebibliography}

\end{document}